\documentclass[twocolumn,letterpaper]{IEEEAerospaceCLS}

\usepackage[]{graphicx}    

\usepackage{amsfonts}
\usepackage{bbm}
\usepackage{ dsfont }
\usepackage{amsmath,amssymb, amsthm}
\usepackage{xcolor}
\usepackage{graphicx}
\usepackage{float}

\def\R{\mathbb{R}}
\def\P{\mathbb{P}}

\begin{document}
\title{Explainability Tools Enabling Deep Learning in Future In-Situ Real-Time Planetary Explorations}

\author{%
Daniel Lundstrom\\ 
University of Southern California\\
3620 S. Vermont Ave., KAP 104\\
Los Angeles, CA 90089\\
lundstro@usc.edu
\and 
Alexander Huyen\\
Jet Propulsion Laboratory\\
California Institute of Technology\\ 
4800 Oak Grove Dr., Pasadena, CA 91109 \\
alexander.l.huyen@jpl.nasa.gov
\and 
Arya Mevada\\
California Institute of Technology\\ 
1200 E. California Blvd. MSC 643, \\Pasadena, CA 91125 \\
amevada@caltech.edu
\and 
Kyongsik Yun\\
Jet Propulsion Laboratory\\
California Institute of Technology\\ 
4800 Oak Grove Dr., Pasadena, CA 91109 \\
kyongsik.yun@jpl.nasa.gov
\and 
Thomas Lu\\
Jet Propulsion Laboratory\\
California Institute of Technology\\ 
4800 Oak Grove Dr., Pasadena, CA 91109 \\
thomas.t.lu@jpl.nasa.gov


\thanks{\footnotesize 978-1-6654-3760-8/22/$\$31.00$ \copyright2022 IEEE}              
}

\maketitle

\thispagestyle{plain}
\pagestyle{plain}

\maketitle

\thispagestyle{plain}
\pagestyle{plain}

\begin{abstract}
Deep learning (DL) has proven to be an effective machine learning and computer vision technique. DL-based image segmentation, object recognition and classification will aid many in-situ Mars rover tasks such as path planning and artifact recognition/extraction. However, most of the Deep Neural Network (DNN) architectures are so complex that they are considered a 'black box'.  In this paper, we used integrated gradients to describe the attributions of each neuron to the output classes.  It provides a set of explainability tools (ET) that opens the black box of a DNN so that the individual contribution of neurons to category classification can be ranked and visualized. The neurons in each dense layer are mapped and ranked by measuring expected contribution of a neuron to a class vote given a true image label. The importance of neurons is prioritized according to their correct or incorrect contribution to the output classes and suppression or bolstering of incorrect classes, weighted by the size of each class. ET provides an interface to prune the network to enhance high-rank neurons and remove low-performing neurons. ET technology will make DNNs smaller and more efficient for implementation in small embedded systems. It also leads to more explainable and testable DNNs that can make systems easier for Validation \& Verification. The goal of ET technology is to enable the adoption of DL in future in-situ planetary exploration missions.
\end{abstract} 

\tableofcontents

\section{Introduction}
\subsection{Background}
In recent years, Deep Learning (DL) has become so popular that it has found applications in many fields \cite{lecun2015deep}, \cite{emmert2020introductory}. In the early 1990s, the neural network architectures were quite simple, mostly consisting of “Shallow Networks” with a few hidden layers. Into the 21st century, the sizes of the neural networks grew rapidly due to the advancement of computer hardware technology, especially Graphical Processing Units (GPU) boards now widely used in accelerating the neural network training and processing. For example, many popular DL models were created in the “ImageNet Large Scale Visual Recognition Challenge” \cite{russakovsky2015imagenet} in order to mimic or even surpass human visual ability. As the accuracy of recognizing objects in 10s of millions of photographs increases, the network size also grows ever larger. Floating point operations (FLOPs) are used to measure the network size, or how many operations are required to run a single instance of a neural network model. AlexNet with 8 layers has approximately 720 million FLOPs, GoogleNet with 22 layers has about 1.5 billion FLOPs, and the ResNet with 152 layers has 11.3 billion FLOPs \cite{yun2019accelerating}, \cite{yun2018deep}. These models were developed due to massive data sets such as ImageNet. However, these DL models have been adopted by many other applications that have limited training and testing data. There could be several major drawbacks of adopting a DL model for a new application: (1) The DL model could overfit the training data resulting in many unused neurons. (2)	The training and processing would require large amounts of memory and the convergence speed would be very slow. (3)	The DL models would become so big and complex that they would become “Black Boxes” to the user. In these situations, the testing and validation of the large models would become very difficult.

In aerospace applications, the embedded processors have limited computing power and memory.  In addition, the validation and verification (V\&V) is critically important to the safety and reliability of the spacecraft systems. The DL models in the current format have become difficult to be applied to the aerospace field.  This paper addresses two issues in the DL models: (1) Explainability of the DL models; and (2) Efficiency of the DL models.

This paper is divided into two parts.  First, The DL models are opened up, the neurons in the convolutional layers and dense layers are mapped and ranked against the training output classes. The contribution of each neuron to the output classes can be explained using this methodology.  Secondly, the ranking of the neurons also aids the pruning of the neural network.  A pruning strategy is developed based on the ranking of the importance of each neuron toward the output classes.   

\subsection{Neural Network Pruning}
Neural network pruning is the process of optimization by removing unimportant connections or nodes from a network inspired by human brain development \cite{seeman1999brain}. Pruning is a natural process that occurs in the human brain between infancy and adulthood. During pruning, the brain removes extra synapses. Synapses are brain structures that allow neurons to transmit electrical or chemical signals to other neurons (i.e., weights in artificial neural networks). 

In artificial neural networks, we first evaluate the importance of neurons. And we remove the least important neurons. We then retrain or fine-tune the network and decide whether to continue or stop pruning depending on the degree of degradation. We evaluate pruning by accuracy, size, and computation time \cite{blalock2020state}. 

The effectiveness of pruning depends on the layer we are targeting to prune. For example, in VGG16, 90\% of the weights are in the fully-connected layers, but these weights account for only 1\% of the total floating point operations \cite{li2016pruning}. Pruning fully-connected layers can significantly reduce model size, but cannot improve processing speed. There are two types of pruning: (1) Unstructured pruning that sets each individual weight to zero. Storage can be saved by using unstructured pruning, but little to no speedup is expected \cite{hoefler2021sparsity}. (2) Structured pruning where we set the entire weight columns to zero. By removing entire neurons, filters, or channels from the target layer, we can expect both storage savings and speed enhancement through dense computation \cite{li2016pruning}. 
The importance ranking of each node in an artificial neural network is based on the L1 norm of each filter weight (sum of absolute vector values) \cite{li2016pruning}. The masking method is to mask the filters represented by neurons (or columns of neurons) one by one, calculate network accuracy, and perform scoring based on the results. Masking can be done for all desired weights in a single step, or it can be scheduled iteratively over multiple steps for parts of the network \cite{castellano1997iterative}.


\section{DNN Explainability Research}


\subsection{Theory}
We develop a method of attributions for a layer of internal neurons in a feed-forward neural network with a Softmax output. The method is based on the Integrated Gradients (IG) method developed by Sundararajan and colleagues \cite{sundararajan2017axiomatic}. We begin by supposing an object recognition feed forward neural network $F: \R^\emph{l} \longrightarrow \R^\emph{n}$, an input image $x$, and a baseline input image $x'$. The baseline suggested in Sundararajan's work is an input representing a lack of any information or features, such as a black image. IG attributions indicate how much each entry of the input image contributes to the difference in output, $F(x) - F(x')$. For a given input entry $x_i$ and output entry $F_j(x)$, the IG attribution is:

\begin{equation}
    A_{ij}(x) = (x_i - x_i') \int_0^1 \frac{\partial F_j(x' + \alpha(x - x') )}{\partial x_i} d\alpha
\end{equation}

The IG can be likened to taking the average sensitivity of an output $F_j$ to local perturbations of the input $x_i$ over all images on the path $x' + \alpha(x - x'), \alpha \in [0,1]$.

In the context of an object recognition model with a softmax output, a positive IG attribution indicates the input contributed to an increase in the category vote, while a negative value indicates the input suppressed the category's vote. The IG method of attributions is shown to satisfy certain desirable qualities, the full of which can be found in the original paper. One noteworthy property is that the sum of attributions for a given input equals the difference in output: $\sum_{i=1}^l A_{i,j}(x,x') = F_j(x') - F_j(x)$. Thus, IG attributions can almost be thought of as divvying up the output difference amongst the inputs. An exception to this conception is that it is possible for an individual attribution to be negative even though the category vote increased from the baseline. This would indicate that the input suppressed the vote, but other inputs contributed to it's increase, thus increasing the vote overall. If this occurs, conservation of the sum could cause the sum of the total positive attributions to exceed the difference in outputs. Another noteworthy property of IG is that if an input does not change from it's baseline, or $x_i = x_i'$, then $A_{ij}(x,x') = 0$. In our context, this will cause dead neurons to have an attribution of zero.

We apply this attribution method to a dense layer of internal neurons in a neural network. Suppose we split the neural network at a dense layer creating two halves, $G$ and $H$. Formally, $F(x) = G(H(x))$ where $H$ and $G$ are two neural network, $H:\R^\emph{l} \longrightarrow \R^ \emph{m}$ and $G:\R ^{\emph{m}} \longrightarrow \R ^\emph{n}$, and the first layer of $G$ is a dense layer. Using the method of IG we can get attributions for the layer of neurons that are the input of $G$. In our current model, we use ReLU activation, so a baseline of 0 is the minimum value H can take and represents the absence of any features in the input of $G$. Because of this, we set the baseline to $0$ and define the attribution of neuron $i$ to output $j$ for image $x$ as
\begin{equation}
    A_{ij}(x) = H_i(x) \int_0^1 \frac{\partial G_j(\alpha H(x) )}{\partial H_i} d\alpha
\end{equation}

Representing our data set as a pair of inputs and labels $(x,y)$, we define the expected attribution (EA) as:
\begin{equation}
    E_{ij} = \mathbb{E} \hspace{1mm} \emph{A}_{\emph{ij}}(\emph{x})
\end{equation}

Naturally, $E_{ij}>0$ for $k=j$ indicates that on average, neuron $i$ bolsters the vote for a category, while $E_{ij}<0$ indicates that the neuron suppresses the category's confidence. This statistic is defined over the whole data set, but it may be of interest to measure the EA of a neuron for a subset of the data. To this end, we define the conditional EA as:
\begin{equation}
    E_{ij}^\emph{k} = \mathbb{E} (\emph{A}_{\emph{ij}} (\emph{x})|\emph{y=k})
\end{equation}

Here $E_{ij}^k$ is the expected attribution of neuron $i$ contributing to the model voting for category $j$, over the subset of images who's true labels are $k$. Naturally, $E_{ij}^k>0$ for $k=j$ indicates that on average, neuron $i$ bolsters the vote for a category when that category is correct, while  $E_{ij}^k <0$ for $k=j$ indicates the opposite. In the other case, $E_{ij}^k < 0$, $j\neq k$ indicates that neuron $i$ suppresses the vote for an incorrect category, while $E_{ij}^k > 0$, $j\neq k$ indicates contribution to an incorrect vote. For computational efficiency, it is only necessary to compute the conditional, as expected attributions may be computed from the conditional by the law of total expectation:

\begin{equation}
    E_{ij} =  \sum_k E_{ij}^k \P(\emph{y}=\emph{k})
\end{equation}

Thus far we have outlined the application of EA to map the contribution of neurons to confidence outputs. In order to rank the overall effectiveness of neurons, we must consider the model's confidence in the correct category. As an example, a neuron that contributes $0.2\%$ to correct classification when the model is $95\%$ confident is less important that a neuron that contributes the same percentage when the model is $50\%$ confident. This consideration is reflected in the formulation of the categorical cross-entropy loss function. If a model $F$ applied to an input $x$ gives a confidence of $p_y$ for the correct category, then the cross entropy loss is given as:

\begin{equation}
    L_{CE}(F,x) = -\text{log} \hspace{1mm} p_y
\end{equation}

We can directly attribute a neuron's contribution to the loss function, as the loss is a direct function of the neuron values. We define the rank of a neuron as the negative of expected attribution to the loss function. We flip the sign because suppressing the loss corresponds with better performance, and corresponds to a negative attribution to loss.

\subsection{Pruning Approach/Method}
Pruning is implemented through deactivating specific neurons in individual layers by masking their outputs. We first train a regular model with a standard layout, and then apply our ranking function to a desired layer. The trained model is split up into two halves, with the split occurring at the layer to be ranked. The first half of the network ends at the ranking layer, with the second half of the network sequentially continuing from that previous layer. Each of these split models are loaded with the identical weights from the full model in respect to its matching layer. Both of these split models are used in conjunction to generate the attributions based on the image data category correctness and ranks every neuron inside a particular layer. This neuron ranking process is done individually for each layer of interest. In our experiment, we applied this ranking method to three layers towards the end of the model - one flatten layer followed by two dense layers. Dropout layers are used after each dense layer but were not ranked or pruned.

When ranking for each of the three layers is finalized, masking layers are added. The full model layout is cloned with the addition of new masking layers after each ranked layer. The same weights from the full model are loaded into the matching layers of the masked model. All masking layers begin with all neurons active, with neurons being deactivated based on the sorted rankings from least to most important. As the number of pruned neurons increases, more higher ranked neurons are deactivated, resulting in a significant decrease in model accuracy.

The pruned network and full network are used to classify the same data set for comparing accuracy and sparsity trade-offs between the models. This method of deactivating neurons does not require refitting of the model. However, refitting the pruned network can increase the model accuracy while still maintaining the same level of neuron sparsity. Specific neurons can be toggled quickly through this method as long as a masking layer is implemented for that layer.

\section{Experimental Results}

\subsection{Training}
In order to test and validate the methods, it was necessary to create a well trained neural network that could differentiate between different classes of images. For this, we referred to a well known image classification network known as AlexNet. Created in 2011, AlexNet has shown very high accuracy in difficult image classification tasks\cite{NIPS2012_4824}.

\begin{figure}[h]
\centering
\includegraphics[width=6cm]{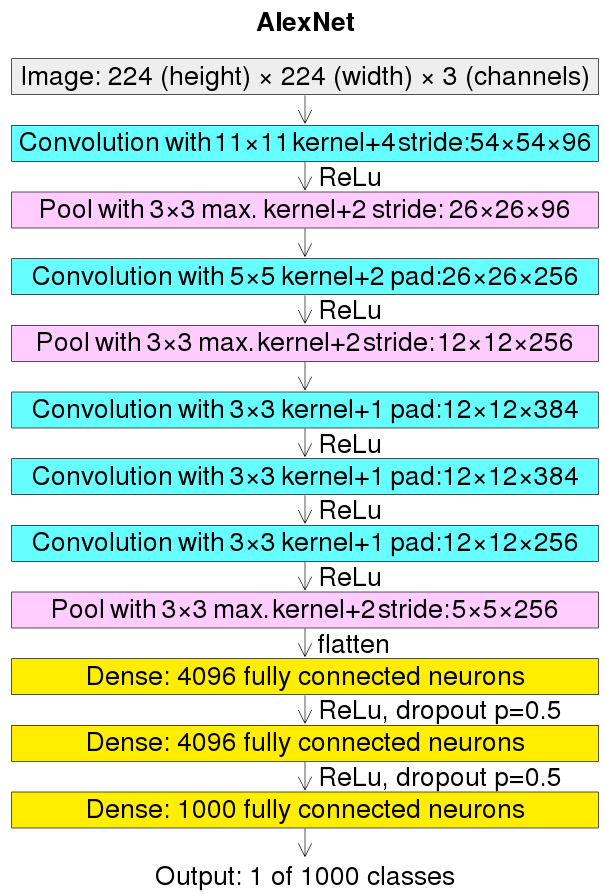}\\
\caption{\textbf{AlexNet Architecture}}
\label{fig:CNNArchitecture}
\end{figure}

The neural network contains 2 Dense layers of 4096 neurons each where most of the parameters are, and a third dense layer for the predictions, so ranking and pruning those are of great importance since those operations are the ones that decide the output. This structure has been modified and perfected and is similar to other image classification networks, such as VGG16/19 and LeNet (which is a predecessor to AlexNet), and as such is a good model to use because any experiments that are tested here can be applied to many other kinds of networks. The full architecture for AlexNet is shown in Fig.~\ref{fig:CNNArchitecture}.

\begin{figure}[H]
\centering
\includegraphics[width=8cm]{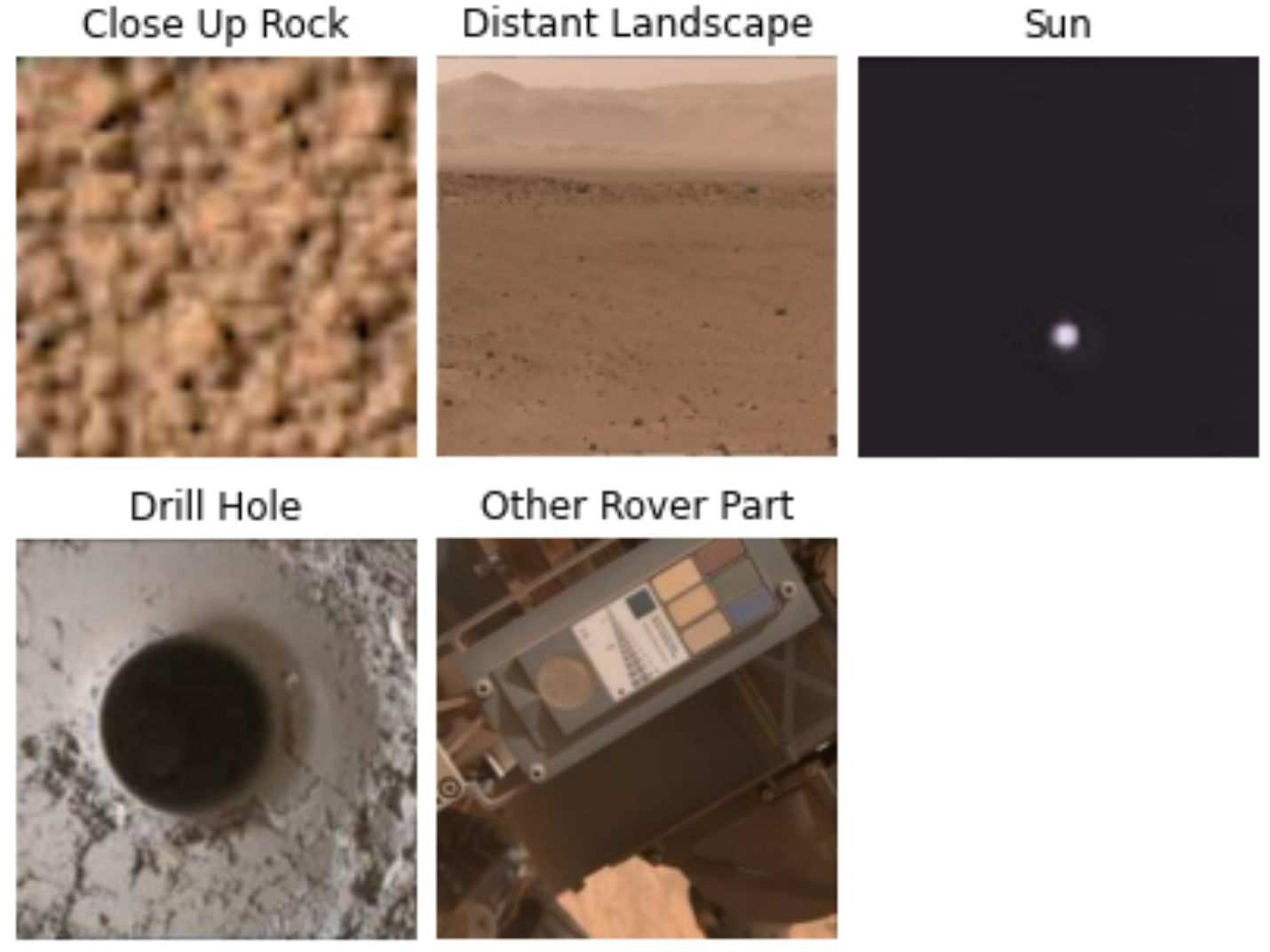}\\
\caption{\textbf{Examples of each of the five classes of images}}
\label{classExamples}
\end{figure}

The network was then trained on a data set of images from the Mars Rover \cite{DBLP:journals/corr/abs-2102-05011}. The images consisted of 5 main categories: close up rock, distant landscape, drill hole, other rover part, sun. There were 19 total classes in the set, however, these 5 classes gave a good mix of distinction (many of the classes are very similar) as well as relevance to the tasks of the rover. The data was then cleaned to remove any augmented or transformed images then randomly shuffled into a 80/10/10 split between train/test/validation. A summary of the dataset distribution is shown in Fig.~\ref{classExamples}. 

\begin{table}[H]
\renewcommand{\arraystretch}{1.3}
\caption{\bf Breakdown of classes in training set}
\label{TrainingSetSplits}
\centering
\begin{tabular}{|c|c|}
\hline
\bfseries Image Class & \bfseries Number Training Set (Percentage) \\
\hline\hline
Close up Rock & 301 (26.6\%) \\
Distant Landscape  & 302 (26.7\%)\\
Drill Hole & 206 (17.8\%)\\
Other Rover Part & 162 (14.3\%)\\
Sun & 161  (14.2\%)\\
\hline
\end{tabular}
\end{table}
The data set consisted of 1132 training images, 164 validation images, and 153 testing images. The set of images was broken up into batches of 32, and a real-time data augmentation function was used to augment the data.  We initialized the network in TensorFlow using their Sequential API, and trained it using their normal training loop. The network was trained for 50 epochs using a constant learning rate of $1 \times 10^{-3}$. After the network was trained, the best iteration of the model was chosen based on its results on the validation set, and then was tested against the test set. The results of these tests are shown in Table \ref{Training Results}. 

\begin{table}[H]
\renewcommand{\arraystretch}{1.3}
\caption{\bf Summary of data set sizes}
\label{Training Results}
\centering
\begin{tabular}{|c|c|c|}
\hline
\bfseries Data Set & \bfseries Number Images  & \bfseries Test Accuracy \\
\hline\hline
Training  & 1132  & 91.6\%\\
Validation & 164  & 85.6\%\\
Testing & 153 & 84.2\% \\
\hline
\end{tabular}
\end{table}


\subsection{Ranking and Expected Attributions Results}

For our experiments, the expected attributions (EAs) and ranking for the input of the three dense layers are calculated. EAs and rankings are generated from training set data and validation tests are employed on the test set. Experiments performed on a single layer were performed on dense layer 2.
\begin{figure}[H]
\centering
\includegraphics[width=8cm]{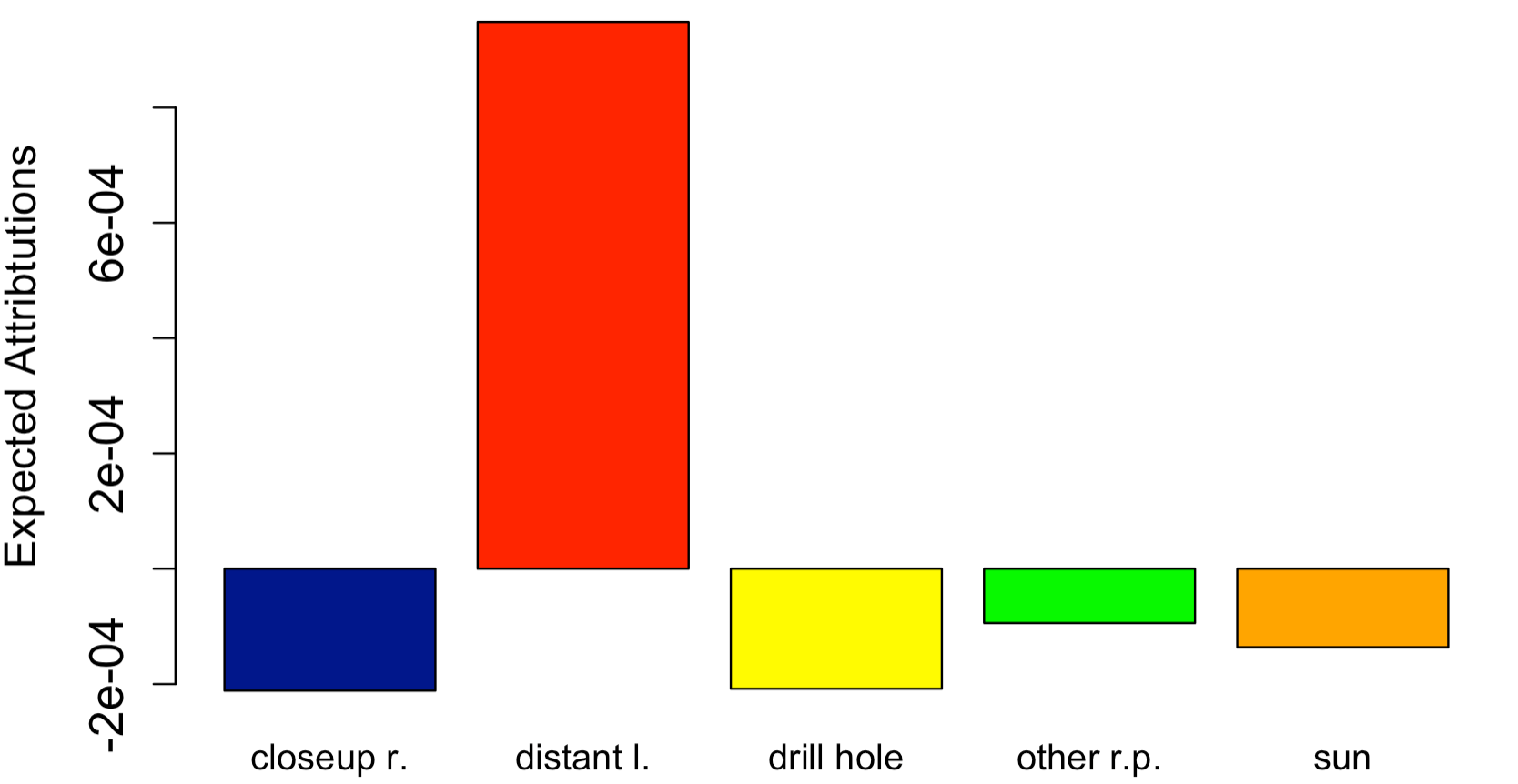}\\
\caption{\textbf{A high-ranked neuron's expected attributions (EAs) to each category.}}
\label{fig:AN}
\end{figure}
Fig.~\ref{fig:AN} is an example of a general expected attribution for a neuron. It shows average contribution to a category vote for the training set. The EA values indicate that the birds-eye trend of the neuron was to boost the confidence of distant landscape and suppress every other category. It did not suppress every other category equally; it appears it suppressed close up rock and drill hole twice as much as other rover part. From this graph, it is hard to tell if the neuron was functioning effectively, since we can't tell if the neuron was boosting distant landscape for those images or another image subset, and similarly for suppression. The y-axis indicates what portion of the confidence the neuron boosted and suppressed. These value may be diluted, since it may be that the neuron acts for certain sub-categories of images strongly, and had zero activity for the majority of images. The next figure gives us more detail.

\begin{figure}[H]
\centering
\includegraphics[width=8cm]{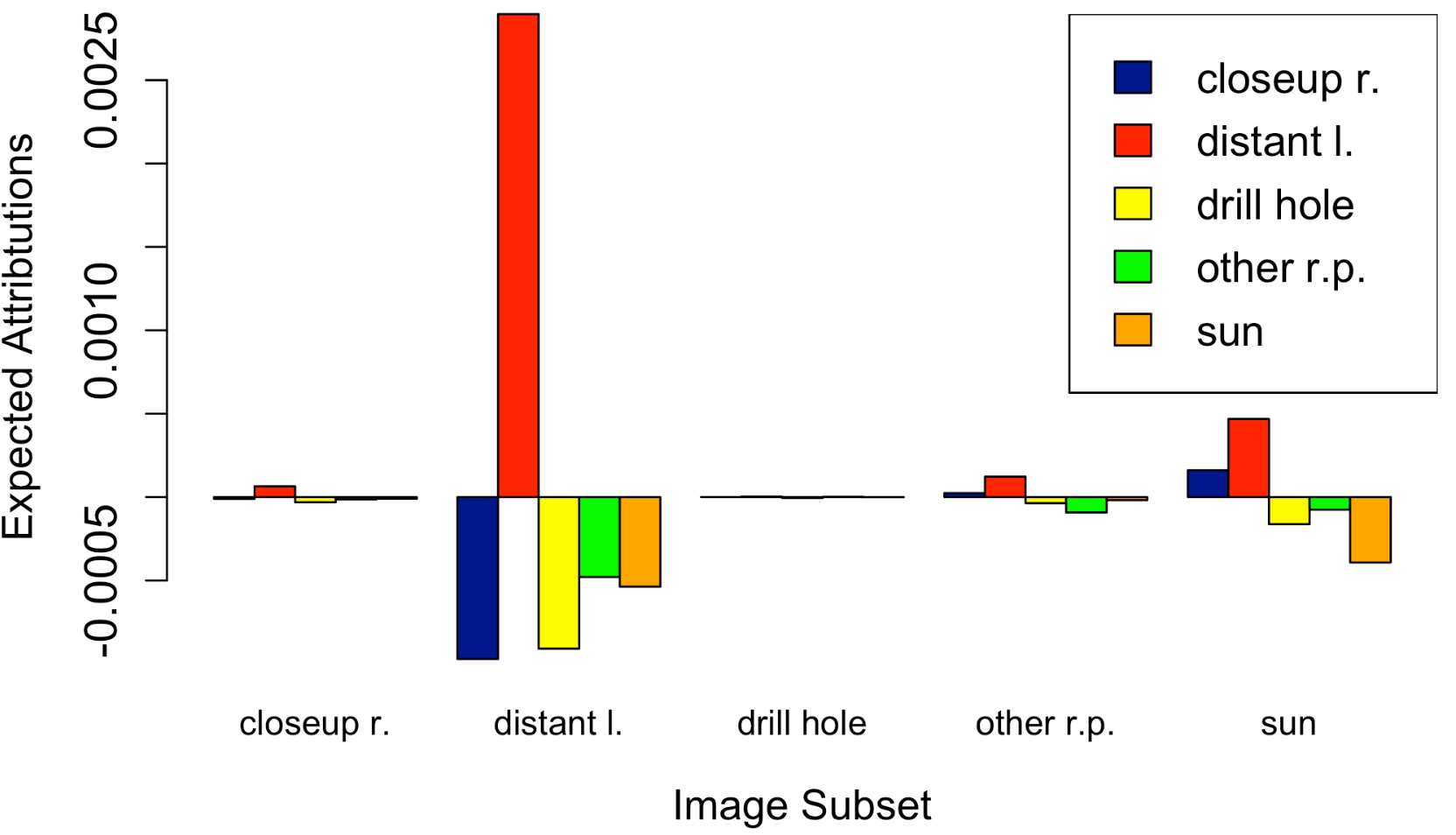}\\
\caption{\textbf{A higher-ranked neuron's expected attributions, broken down by image subcategory.}}
\label{fig:HN}
\end{figure}
\begin{figure}[H]
\centering
\includegraphics[width=8cm]{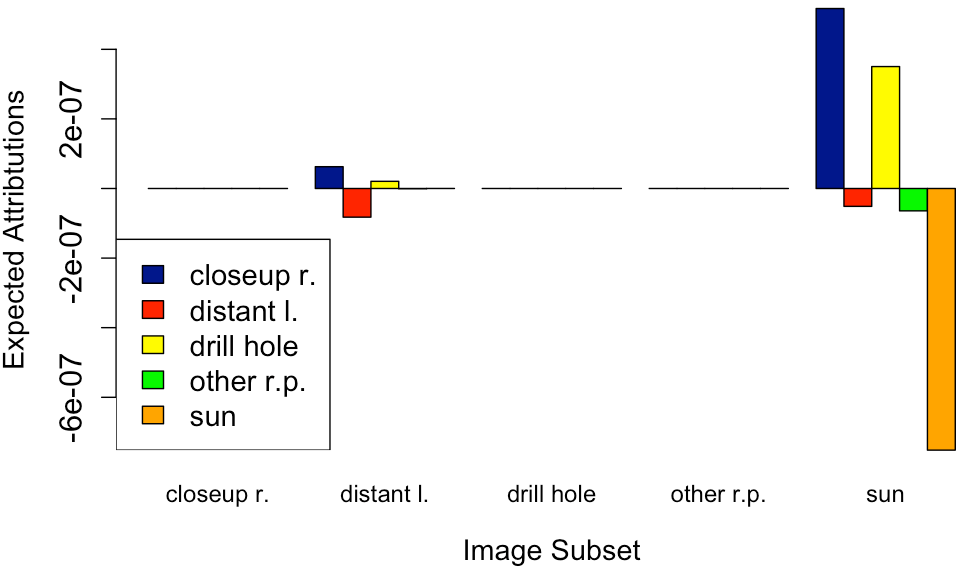}\\
\caption{\textbf{A low ranked neuron's expected attributions, broken down by image subcategory.}}
\label{fig:LN}
\end{figure}

Fig.~\ref{fig:HN} is a more detailed view of the same neuron from Fig~\ref{fig:HN}, with results separated into separate image sets. EA indicates that when the images were distant landscape, the neuron promoted votes to the true label and suppressed every other label, suppressing closeup rock and drill hole twice as much as the others. In the image subcategory of sun, it generally boosted distant landscape and some closeup rock, while it suppressed sun, the correct label. From the graph, it would make sense that this is a higher ranked neuron, since it significantly boosted distant landscape when it was the true label, and didn't harm any other predictions except the sun category, which it did very mildly by comparison. Fig.~\ref{fig:LN} shows a low ranked neuron. This neuron suppressed the correct label when it was active, particularly the sun category.

\begin{figure}[H]
\centering
\includegraphics[width=8cm]{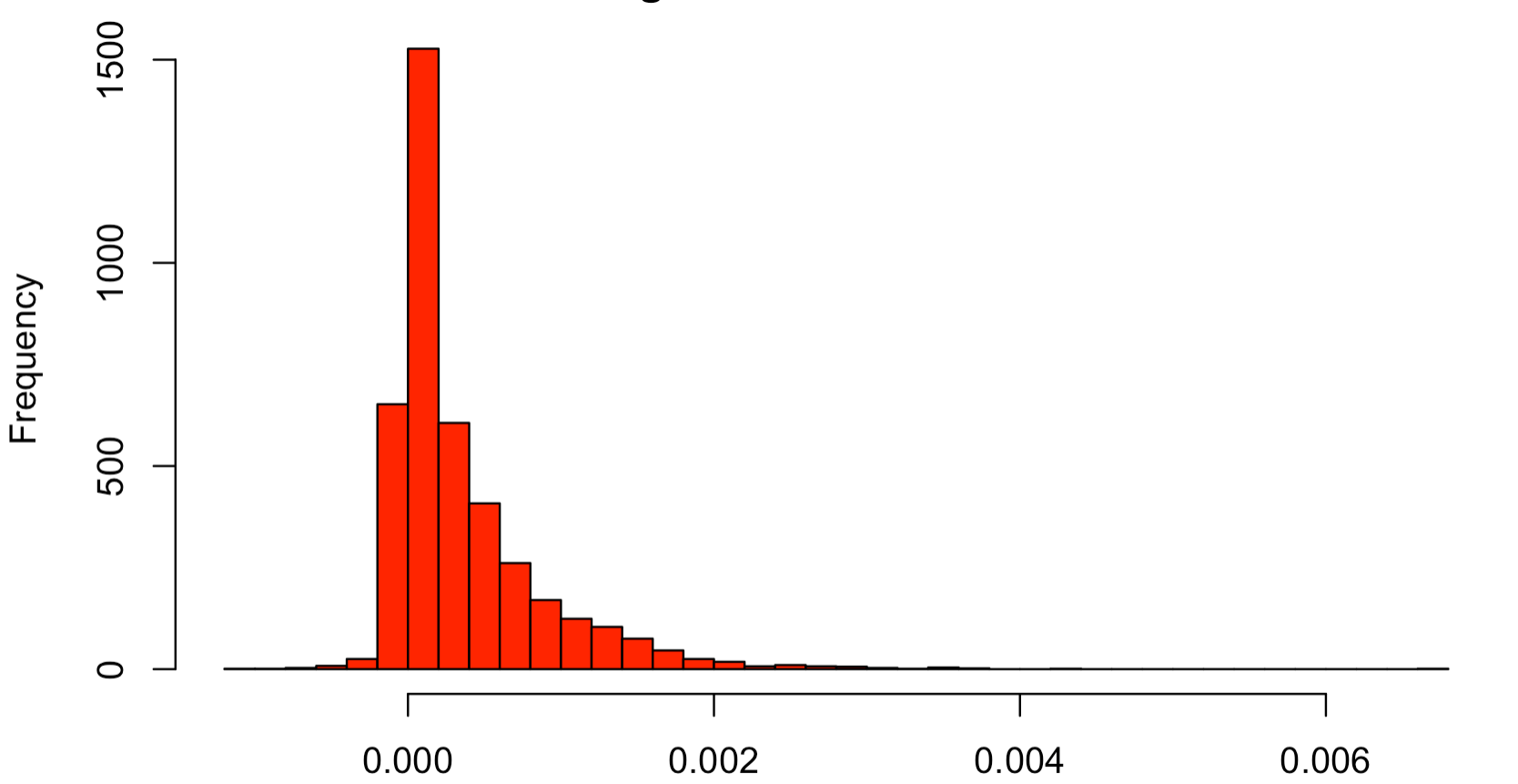}\\
\caption{\textbf{This figure shows the distribution of ranks among all live neurons. Most neurons had a rank near zero, and significant positive rankings vastly outnumbered significant negative rankings. }}
\label{fig:neuron_ranks}
\end{figure}

The distribution across all neurons, both for rankings and conditional EAs, had a similar shape. Fig.~\ref{fig:neuron_ranks} gives a histogram of the ranks, where we see the vast majority were positive, with a grouping around 0 and a right tail. Note that there were a few highly ranked neurons comprising the tail of a distribution. The motivation behind pruning is to cut all but these highly ranked neurons and maintain most of the model's performance.

\begin{figure}[H]
\centering
\includegraphics[width=8cm]{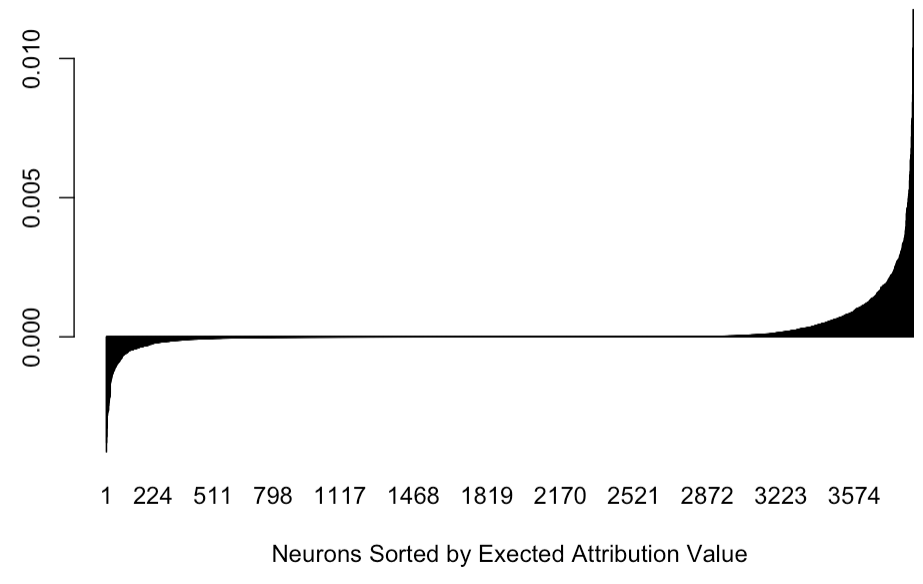}\\
\caption{\textbf{Neuron EA for sun, when the data is restricted to the Sun image subset. This distribution is typical of all sorted expected attributions and sorted rankings.}}
\label{fig:neuron_ranks_cat.png}n
\end{figure}

\begin{figure}[H]
\centering
\includegraphics[width=8cm]{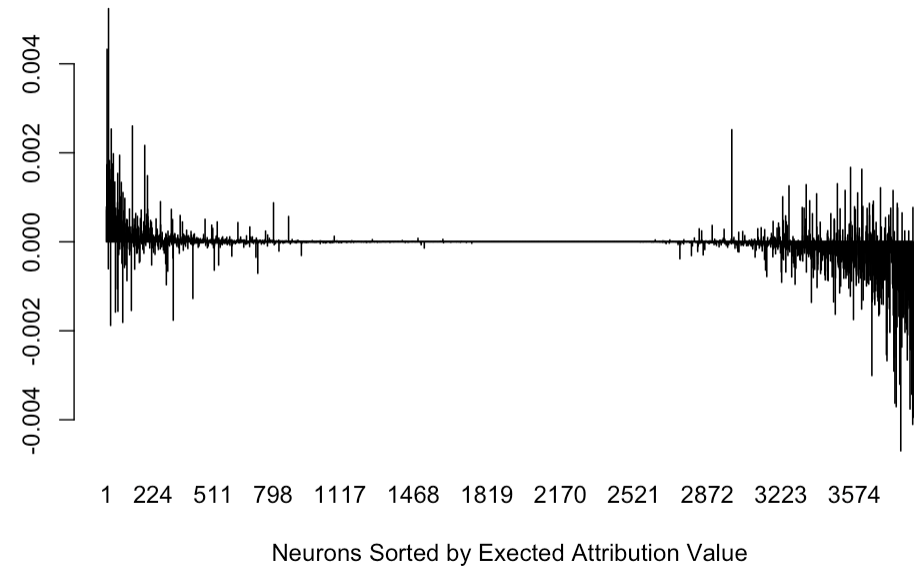}\\
\caption{\textbf{Neuron EAs for drill hole, when the data is restricted to the Sun image subset. On the Sun data subset, drill hole had fewer positive attributions and more negative attributions.}}
\label{fig:neuron_ranks_off_cat.png}
\end{figure}

Fig.~\ref{fig:neuron_ranks_cat.png} gives an alternative view to this common distribution, except the EA values were sorted according to contribution to sun and calculated on the set of sun images. In~Fig.~\ref{fig:neuron_ranks_off_cat.png} the same ordering and image set were kept, but the attribution category was changed to a different class. These figures combined show that, for sun images, most neurons that boosted sun didn't boost, or even suppress, drill hole, and many neurons that suppressed sun boosted drill hole.

We observed a general trend, which is that EA value curves were generally larger in magnitude when attributing to categories of the same image subset, and smaller for off-category attributions (magnitude of Fig.~\ref{fig:neuron_ranks_cat.png} vs Fig.~\ref{fig:neuron_ranks_off_cat.png}). This is a sign of a well functioning network, and is due to the distribution of confidences at the baseline. Classifying a baseline for most models should give roughly equal confidence to each category, while a correct classification for an image will boost one category up to nearly one and drop the other categories to zero. Thus, for correct classifications, the correct category generally increases by several times the quantity that an incorrect category drops. So in a well functioning network, boosting occurs on a larger magnitude than suppressing, and those attributions should be larger.

\begin{figure}[H]
\centering
\includegraphics[width=8cm]{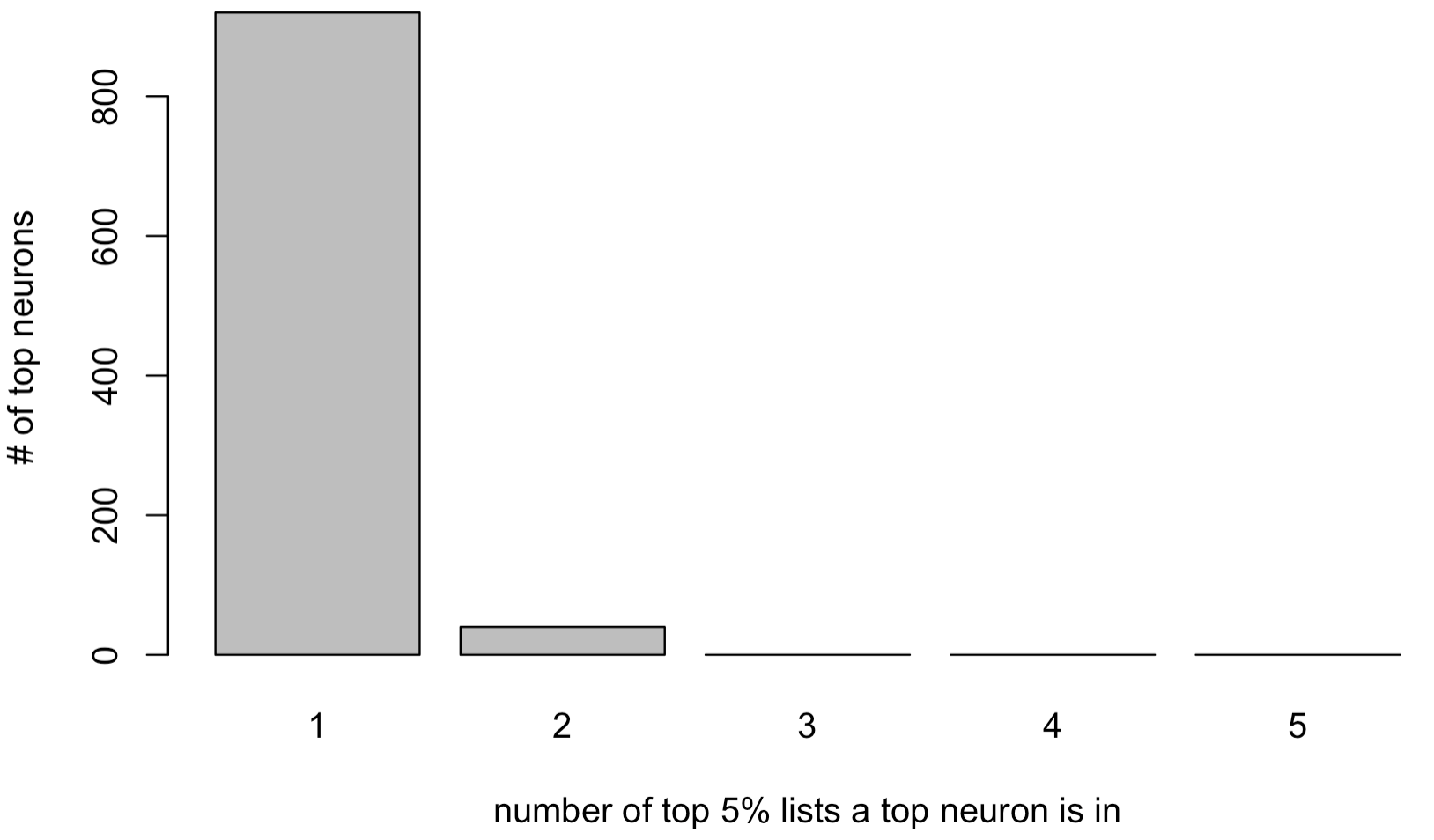}\\
\caption{\textbf{For each data subset, neuron's in the top 5\% rankings were listed. Then for each neuron, the number of lists a neuron was in were counted. Most top-neurons were only in one list.}}
\label{fig:duplicate_tops.png}
\end{figure}

It is possible to create ranking for a particular category by restricting the image set to a particular category. This gives a list of neurons based solely on their performance on a given class of images. We tested for overlap of top-ranked images for each subset of images (Fig.~\ref{fig:duplicate_tops.png}), and found that overlap was minimal, indicating that most highly-effective neurons functioned effectively in only one category, and there are no neurons central to distinguishing between all categories.

\begin{figure}[H]
\centering
\includegraphics[width=8cm]{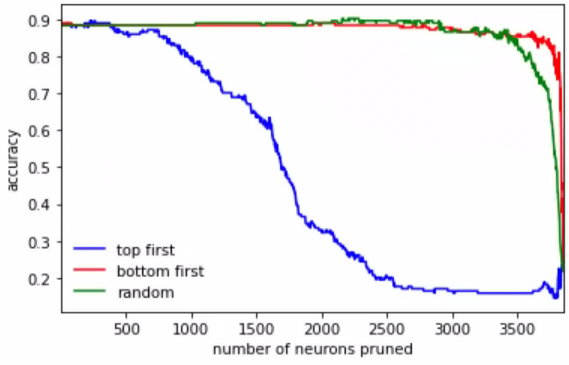}\\
\caption{\textbf{Accuracy vs number of neurons pruned, according to three methods of pruning: prune highest ranked first (top first), and lowest ranked first (bottom first), and random pruning (random).}}
\label{fig:total_loss}
\end{figure}

Internal neuron attributions were validated using pruning experiments on the rankings and expected attributions. First, dead neurons that never activate during the training set were removed, leaving 3894 neurons. These neurons are also neurons with total rank 0. The neurons were then sorted based on rank and pruned three different ways: best neurons first, worst neurons first, and random ordering. The neurons were progressively masked one at a time by zeroing-out their values using a mask. Accuracy on a withheld test set is reported. Fig~\ref{fig:total_loss} shows that if the highest ranked neurons are pruned first, the accuracy quickly drops, while pruning random and lowest first do much less damage to accuracy. The figure also shows that if the worst neurons are pruned first, more neurons can be pruned before the accuracy drops. The delay in accuracy drop when pruning highest ranked first can be attributed to robustness of the model, rather than to an insignificance of high-ranked neurons. This is evidenced by the fact that when all but the highest ranked neurons are pruned (end of bottom first), the accuracy remains high, indicating that the highest ranked neurons are very effective.

\begin{figure}[H]
\centering
\includegraphics[width=8cm]{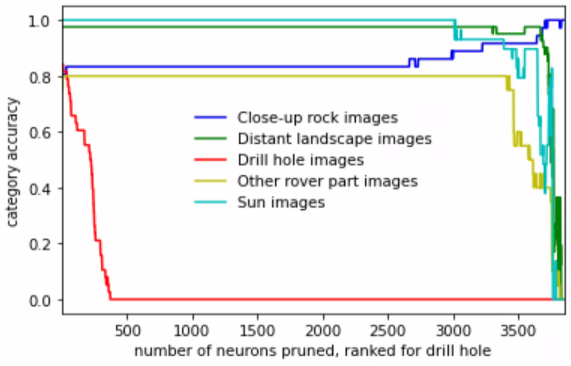}\\
\caption{\textbf{Accuracy when progressively pruning highest attributed neurons for category ``Drill Hole". Each loss is for a different image subset.}}
\label{fig:category_loss}
\end{figure}

Expected attributions were validated by pruning based on EA values. The neurons were sorted based on their EA for a particular category, then pruned from the top down. While pruning, the accuracy was measured for the five different image subsets. Pruning high EA neurons of a particular category effectively reduced the recognition accuracy of that image subset with little effect on other image subsets. This means that each image subset has its own highest performing neuron. Figure ~\ref{fig:category_loss} shows that after pruning the top 400 neurons that EA indicated boosted drill hole, the accuracy on the drill hole image subset is zero while each other categories' performance is unaffected. At around 3500 neurons, the accuracy for the other rover parts data subset is starting to decline, but other categories are not. This suggests that, on the ordered list of neurons boosting drill hole, the neurons that are important to other rover part appear near the last quarter of the list. The fact that the other categories' accuracy are still high indicates that their important neurons are even lower on the list. At the end of the pruning, the accuracy to every category is 0 except close-up rock. This is because the model is no longer varying with the input image, but invariably predicting close-up rock.

\subsection{Pruning Results}

\begin{figure}[H]
\centering
\includegraphics[width=8cm]{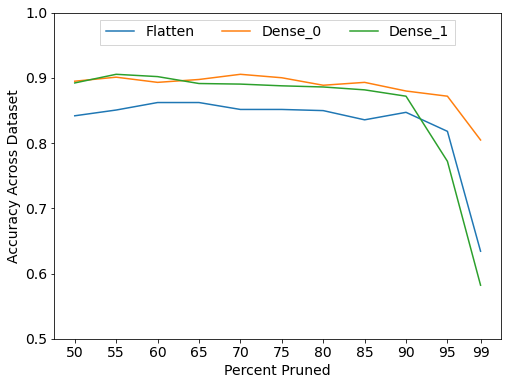}
\caption{\textbf{Model accuracy after pruning, before retraining, for each layer.}}
\label{fig:individual_pruneresults}
\end{figure}

\begin{figure}[H]
\centering
\includegraphics[width=8cm]{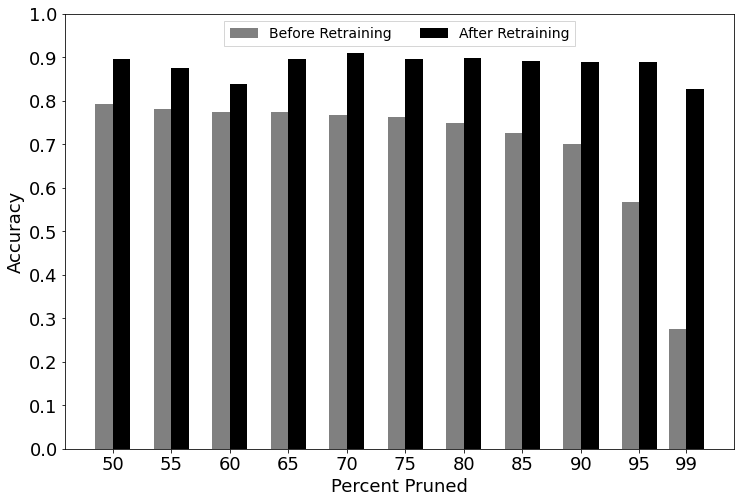}
\caption{\textbf{Ranking and pruning three layers together - Flatten/Dense1/Dense2 layers with and without retraining.}}
\label{fig:layer9_10_12_pruneresults}
\end{figure}

Results from the effects of our ranking and pruning methodology on model accuracy were demonstrated in Figures~\ref{fig:individual_pruneresults} and \ref{fig:layer9_10_12_pruneresults}. Initial results were shown for ranking and pruning individual layers separately. The accuracy was calculated using the training set. Pruning the dense layers resulted in a large reduction in model accuracy on the standard data set. Ranking and pruning each layer independently helped analyze the effectiveness of the ranking method on different types of layers.

A comparison when pruning the multiple layers together and testing the model's accuracy without retraining and after retraining can be seen in Figure \ref{fig:layer9_10_12_pruneresults}. Pruning all three layers (one flatten, two dense) resulted in an expected drop in accuracy with a similar trend seen when pruning the individual layers independently. The model accuracy was shown to improve or recover when the model was retrained after pruning. A significant recovery in model accuracy was seen when pruning a majority of a layer's neurons and retraining with the same amount of pruning. 95\% of neurons were deactivated in each of the three layers, reducing the final model accuracy to 56.7\%. Retraining the 95\% pruned model improved the accuracy to 88.9\%, with each pruning mask layer still left intact and 95\% of neurons remaining deactivated in each of the three layers.

We used the ranking methodology to prune the model and retrain it to see that accuracy is restored. This shows the potential to take advantage of the reduction in the number of parameters while maintaining high model accuracy. Table \ref{table:cutparameters} shows the amount of parameters reduced in reference to the pruned layers seen previously. The accuracy trade-off when pruning large amounts of neurons can be minimized by retraining the pruned model. This process results in a 93.3\% reduction in the total number of parameters pruned across the entire network, while achieving a model accuracy of 95.3\% on the standard test set.

\begin{table}[h]
\renewcommand{\arraystretch}{1.3}
\caption{\bf Reduction of Parameters based on Pruning Percentage. The full network contains 58,301,829 parameters before any pruning.}
\label{table:cutparameters}
\centering
\begin{tabular}{|c|c|c|}
\hline
\bfseries Masking \% & \bfseries Total Parameters Cut  & \bfseries Total \% Pruned \\
\hline\hline
   50\% &   40,908,800 & 70.2\% \\
   55\% &   43,500,216 & 74.6\% \\
   60\% &   45,819,003 & 78.6\% \\
   65\% &   47,865,160 & 82.1\% \\
   70\% &   49,638,687 & 85.1\% \\
   75\% &   51,139,584 & 87.7\% \\
   80\% &   52,367,852 & 89.8\% \\
   85\% &   53,323,489 & 91.5\% \\
   90\% &   54,006,497 & 92.6\% \\
   95\% &   54,416,876 & 93.3\% \\
\hline
\end{tabular}
{\bf }
\end{table}

\section{Analysis and Discussion}

A deep neural network (DNN) has many layers of interconnected neurons. Training a DNN may require a large amount of training data that is difficult to obtain in reality. It is computationally expensive and requires parallel processors, such as graphical processing units (GPUs), to perform the computation. It is also difficult to test, validate and verify (V\&V), and its unpredictability makes it difficult to apply to space exploration. In this study, as a result of classifying images received from the Mars rover using AlexNet, a relatively small deep neural network, it was found that there were many unused neurons. By analyzing the attributions of the neurons in the dense layers, we can rank the neurons based on the values of each neuron's contribution to the output classes.  From the attribution and ranking plots, we can learn the importance of each neuron to the input and output classes. It helps to describe the function of each neuron by quantifying its contribution to the output class. The ranking of the neurons also helps the pruning process. In this study, we found that pruning the network sequentially, starting with lower-ranking neurons, resulted in lower loss of output accuracy than higher-ranking or random pruning. After we removed 95\% of neurons from the network, we observed that the network still worked well. By removing 95\% of neurons, the network became much smaller, from more than 58 million to 4 million parameters. The overall accuracy was 91.6\% before pruning. After 50\% pruning, the overall accuracy was 79.3\%, and after retraining, the overall accuracy was 90.0\%, which was very close to the original accuracy. After 95\% pruning, the overall accuracy was 56.7\%. After retraining, the overall accuracy was restored to 88.9\%, which was also very close to original accuracy.  On smaller networks, it is easier for the training algorithm to converge to a better state. Smaller networks run faster on embedded processors. It also makes testing and validation easier.

\section{Conclusions}

In conclusion, we have shown that the neurons in a deep neural network can be ranked based on their attributions to the input and output classes. By mapping and ranking all neurons in a deep neural network, we can better explain the reasons behind the DNN output class and values. The ranking gives an added benefit in that it can guide the pruning of the network. In the example of classifying objects in the Mars Rover image set using the AlexNet model, we have shown that we can remove 95\% of neurons in dense layers from a network without significant loss of classification accuracy. In fact, by retraining a smaller network after deep pruning, the accuracy can be higher than the original accuracy. The network becomes smaller, faster, and easier to test and validate.


\acknowledgments
The research described in this paper was carried out at the Jet Propulsion Laboratory, California Institute of Technology under a contract with the National Aeronautics and Space Administration (NASA).

\bibliographystyle{IEEEtran}
\bibliography{main}





\thebiography
\begin{biographywithpic}
{Daniel Lundstrom}{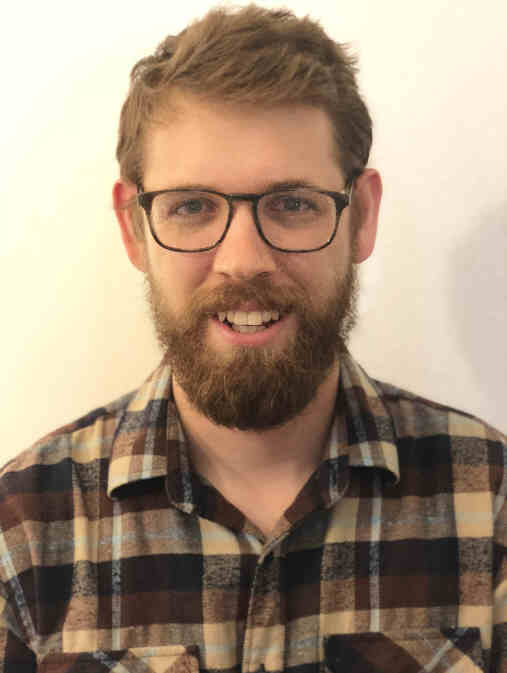}
is an Applied Mathematics Ph.D. student at the University of Southern California. His thesis focus in on explainable artificial intelligence. He has done recent explainable AI research as an intern for Jet Propulsion Laboratory, California Institute of Technology, where he worked with the Department of Transportation to provide increased explainability for AI systems as part of a project to help first responder safety.
\end{biographywithpic} 

\begin{biographywithpic}
{Alexander Huyen}{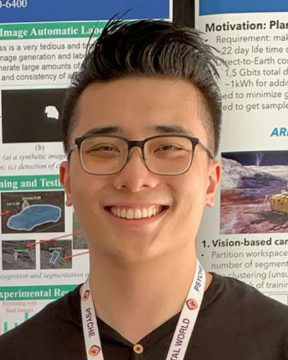} is a software engineer at Jet Propulsion Laboratory, California Institute of Technology. He received his B.S. in Computer Science from California State Polytechnic University, Pomona. His undergraduate course of study involved artificial intelligence, cryptography and information security, computer architecture, and statistics for computer scientists. At JPL, Alex develops engineering applications and conducts machine learning computer vision research. Outside of research, Alex was part of the Simulation Support Equipment software team and Command Data Handling Integration \& Testing team for the Psyche spacecraft under NASA’s Discovery Program. He helped teams develop engineering applications for testing flight software, environment simulators, and instrument telemetry and command Ground Data Systems (GDS).
\end{biographywithpic} 

\begin{biographywithpic}
{Arya Mevada}{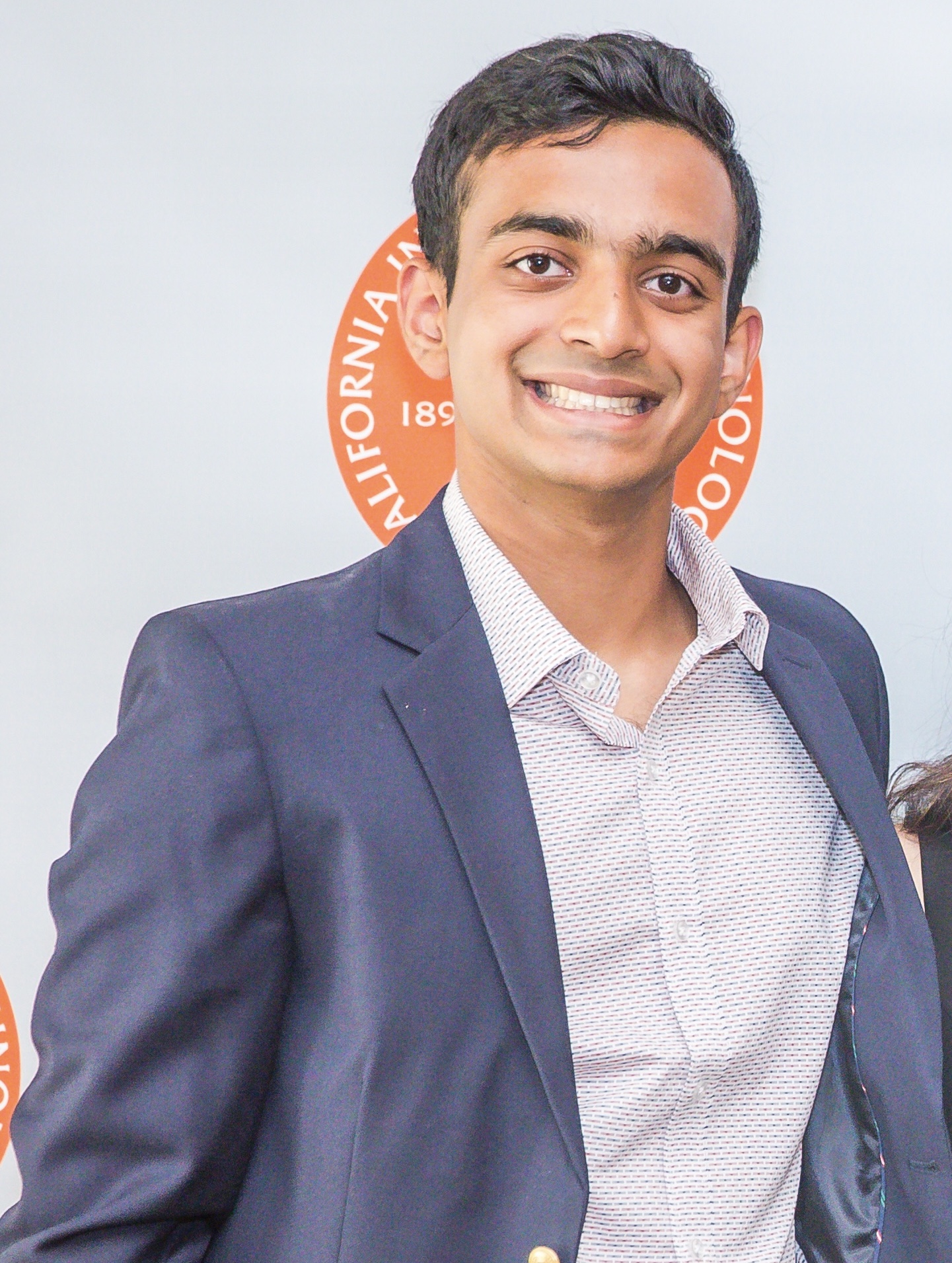} is an 3rd year undergraduate student studying Computer Science at Caltech, with a focus on Machine Learning and Artificial Intelligence. His coursework has been focused on machine learning, statistics, data science and economics with some work in computing and algorithmic theory. He has previous experience working for the Reisman Lab at Caltech, designing and implementing a Computer Vision program to extract chemical reaction data from written literature, specifically looking at determining the direction of reaction flow and ordering of reagents. At the Jet Propulsion Laboratory, he performed research into XAI to help explain different computer vision networks. Outside of the classroom, Arya is a pitcher for Caltech's NCAA Baseball team. 
\end{biographywithpic} 

\begin{biographywithpic}
{Kyongsik Yun}{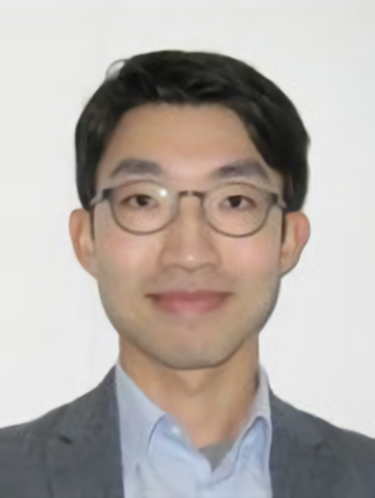} is a technologist at the Jet Propulsion Laboratory, California Institute of Technology, and a senior member of the American Institute of Aeronautics and Astronautics (AIAA). His research focuses on building brain-inspired technologies and systems, including deep learning computer vision, natural language processing, and multivariate time series models. He received the JPL Explorer Award (2019) for scientific and technical excellence in machine learning applications. Kyongsik received his B.S. and Ph.D. in Bioengineering from the Korea Advanced Institute of Science and Technology (KAIST).
\end{biographywithpic}

\begin{biographywithpic}
{Thomas Lu}{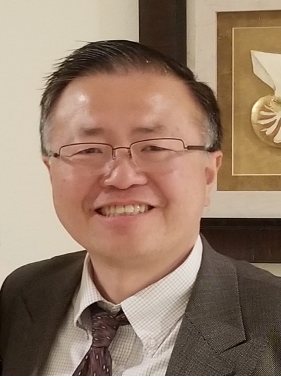} is a Senior Researcher at NASA Jet Propulsion Laboratory, California Institute of Technology.  His research focus has been in data analysis, artificial intelligence (AI), neural network architecture, deep learning, multispectral imaging and computer vision areas. He has served as an organizer and organizing committee member of SPIE conferences, edited a book “Advances in Pattern Recognition Research”, contributed 3 book chapters, co-authored over 70 journal and conference papers,  co-invented 6 patents.    Thomas received his Ph.D. degree in Electrical Engineering from the Pennsylvania State University.
\end{biographywithpic}

\end{document}